\documentclass{article}
\usepackage{ijcai21}

\usepackage{times}
\usepackage{soul}
\usepackage{url}
\usepackage[hidelinks]{hyperref}
\usepackage[utf8]{inputenc}
\usepackage[small]{caption}
\usepackage{graphicx}
\usepackage{amsmath}
\usepackage{amsthm}
\usepackage{booktabs}
\usepackage{algorithm}
\usepackage{algorithmic}
\usepackage{amssymb}
\usepackage{pifont}
\usepackage{times}
\usepackage{epsfig}
\usepackage{amssymb}
\usepackage{amsfonts}
\usepackage{subfigure}
\usepackage{algorithmic}
\usepackage{textcomp}
\usepackage{threeparttable}
\usepackage{multicol}
\usepackage{multirow}
\usepackage{xspace}
\usepackage{bm}
\usepackage{balance}
\usepackage{float}
\usepackage{color}
\usepackage{ragged2e}
\usepackage{amsthm,amsmath,amssymb}
\usepackage{mathrsfs}

\definecolor{qiu}{RGB}{250,10,10}

\def\ie{\emph{i.e.,~}}
\def\eg{\emph{e.g.,~}}

\def\wrt{w.r.t.~}

\def\mata{\textcolor{black}}
\def\ournet{CPGA}
\definecolor{fan}{RGB}{181,68,52}

\newcommand{\cmark}{\ding{51}}%
\newcommand{\xmark}{\ding{55}}%

\title{Source-free Domain Adaptation \\via Avatar Prototype Generation and Adaptation}

\author{
Zhen Qiu$^{1,4}$\thanks{Authors contributed equally.} \and
Yifan Zhang$^{2*}$\and
Hongbin Lin$^{1*}$\and
Shuaicheng Niu$^1$\and
\\
Yanxia Liu$^1$\and
Qing Du$^{1\dag}$\And
Mingkui Tan$^{1,3,4}$\thanks{Corresponding author.}\\
\affiliations
$^1$School of Software Engineering, South China University of Technology\\
$^2$School of Computing, National University of Singapore\\
$^3$Key Laboratory of Big Data and Intelligent Robot, Ministry of Education\\
$^4$Pazhou Laboratory\\
}

\begin{document}

\maketitle

\begin{abstract}

We study a practical domain adaptation task, called source-free unsupervised domain adaptation (UDA) problem, in which we cannot access source domain data due to data privacy issues but only a pre-trained source model and unlabeled target data are available.  This task, however, is very difficult due to one key challenge: the lack of source data and target domain labels makes model adaptation very challenging. To address this, we propose to mine the hidden knowledge in the source model and exploit it to generate
source avatar prototypes (\ie representative features for each source class) as well as target pseudo labels for domain alignment. To this end, we propose a Contrastive Prototype Generation and Adaptation (CPGA) method.
Specifically, CPGA consists of two stages: (1) prototype generation: by exploring the classification boundary information of the source model, we train a prototype generator to generate avatar prototypes via contrastive learning. (2) prototype adaptation: based on the generated source prototypes and target pseudo labels, we develop a new robust contrastive prototype adaptation strategy to align each pseudo-labeled target data to the corresponding source prototypes. Extensive experiments on three UDA benchmark datasets demonstrate the effectiveness and superiority of the proposed method.
\end{abstract}



\section{Introduction}
Unsupervised domain adaptation (UDA) has achieved remarkable success in many applications, such as image classification and semantic segmentation~\cite{Yan2017LearningDC,Liang2019DistantSC,tang2020unsupervised,Zhang2020CollaborativeUD}. 
The goal of UDA is to leverage a label-rich source domain to improve the model performance on an unlabeled target domain, which bypasses the dependence on laborious target data annotation.
Generally, UDA methods can be divided into two categories, \ie data-level UDA and feature-level UDA. Data-level methods~\cite{Sankaranarayanan2018GenerateTA,Hoffman2018CyCADACA} attempt to mitigate domain shifts by image transformation between domains via generative adversarial networks~\cite{goodfellow2014generative}. 
By contrast, feature-level methods~\cite{ganin2015unsupervised,Wei2016DeepNF} focus on alleviating domain discrepancies by learning domain-invariant feature representations. In real-world applications, however, one may only access a source trained model instead of source data due to the law of privacy protection. As a result, many existing UDA methods are incapable due to the lack of source data. Therefore, this paper considers a more practical task, called source-free UDA~\cite{liang2020shot,Li2020ModelAU}, which seeks to adapt a well-trained source model to a target domain without using any source data.
 
Due to the absence of source data as well as target domain labels, it is difficult to estimate the source domain distribution and exploit target class information for alleviating domain discrepancy as previous UDA methods do. Such a dilemma makes source-free UDA very challenging. To solve this task, existing source-free UDA methods seek to refine the source model either by generating target-style images (\eg MA~\cite{Li2020ModelAU}) or by pseudo-labeling target data (\eg SHOT~\cite{liang2020shot}). However, directly generating images from the source model can be very difficult and pseudo-labeling may lead to wrong labels due to domain shifts, both of which compromise the training procedure.



To handle the absence of source data, our motivation is to mine the hidden knowledge in the source model. By exploring the source model, we seek to generate feature prototypes of each source class and target pseudo labels for domain alignment.
To this end, we propose a new Contrastive Prototype Generation and Adaptation (\ournet) method. Specifically, \ournet~contains two stages:
(1) Prototype generation: by exploring the classification boundary information in the source classifier, we train a prototype generator to generate source prototypes based on contrastive learning.
(2) Prototype adaptation: to mitigate domain discrepancies, based on the generated feature prototypes and target pseudo labels, we develop a new contrastive prototype adaptation strategy to align each pseudo-labeled target data to the source prototype with the same class. To alleviate label noise, we enhance the alignment via confidence reweighting and 
early learning regularization.
Meanwhile, we further boost the alignment via feature clustering to make the target features more compact.  
In this way, we are able to well adapt the source-trained model to the unlabeled target domain even without any source data.
 
The contributions of this paper are summarized as follows:
\begin{itemize}
    \item In \ournet, we propose a contrastive prototype generation strategy for source-free UDA. Such a strategy can generate representative (\ie intra-class compact and inter-class separated) avatar feature prototypes for each class. The generated prototypes can be applied to help conventional UDA methods to handle source-free UDA.
    \item In \ournet, we also propose a robust contrastive prototype adaptation strategy for source-free UDA. Such a strategy can align each pseudo-labeled target data to the corresponding source prototype and meanwhile alleviate the issue of pseudo label noise.
    \item Extensive experiments on three domain adaptation benchmark datasets demonstrate the effectiveness and superiority of the proposed method.
\end{itemize}



\section{Related Work}
\paragraph{Unsupervised Domain Adaptation (UDA).}
UDA 
has been widely studied in recent years~\cite{tang2020unsupervised,jin2020minimum}. Most existing methods alleviate the domain discrepancy either by adding adaptation layers to match high-order moments of distributions, \eg DDC~\cite{Tzeng2014DeepDC}, 
or by devising a domain discriminator to learn domain-invariant features in an adversarial manner, \eg DANN~\cite{ganin2015unsupervised} and MCD~\cite{saito2018maximum}. 
Recently, prototypical methods and contrastive learning has been introduced to UDA. For instance, TPN~\cite{pan2019transferrable} and PAL~\cite{hu2020panda} attempts to align the source and target domains based on the learned prototypical feature representations. Besides, CAN~\cite{Kang2019ContrastiveAN} and CoSCA~\cite{Dai_2020_ACCV} leverages contrastive learning to explicitly minimize intra-class distance and maximize inter-class distance in terms of both intra-domain and inter-domain. However, the source data may be unavailable in practice due to privacy issues, making these methods incapable.

\paragraph{Source-free Unsupervised Domain Adaptation.}
Source-free UDA~\cite{kim2020progressive} aims to adapt the source model to an unlabeled target domain without using the source data. Existing methods seek to refine the source model either by pseudo-labeling (\eg SHOT~\cite{liang2020shot}) or by generating target-style images (\eg MA~\cite{Li2020ModelAU}). However, due to the domain discrepancy, the pseudo labels can be noisy, which is ignored by  SHOT. Besides, directly generating target-style images from the source model can be very difficult due to training difficulties of GANs. Very recently, BAIT~\cite{Yang2020UnsupervisedDA} proposes to use the source classifier as source anchors and use them for domain alignment.
However, BAIT requires dividing target data into certain and uncertain sets via prediction entropy of source classifier, which may lead to wrong division due to domain shifts.

Compared with the above methods,  we propose to generate source feature prototypes for each class instead of directly generating images. Besides, we alleviate the negative transfer brought by noisy pseudo labels through confidence reweighting and regularization.


\begin{figure*}[h]
\centering
\includegraphics[width=15.8cm]{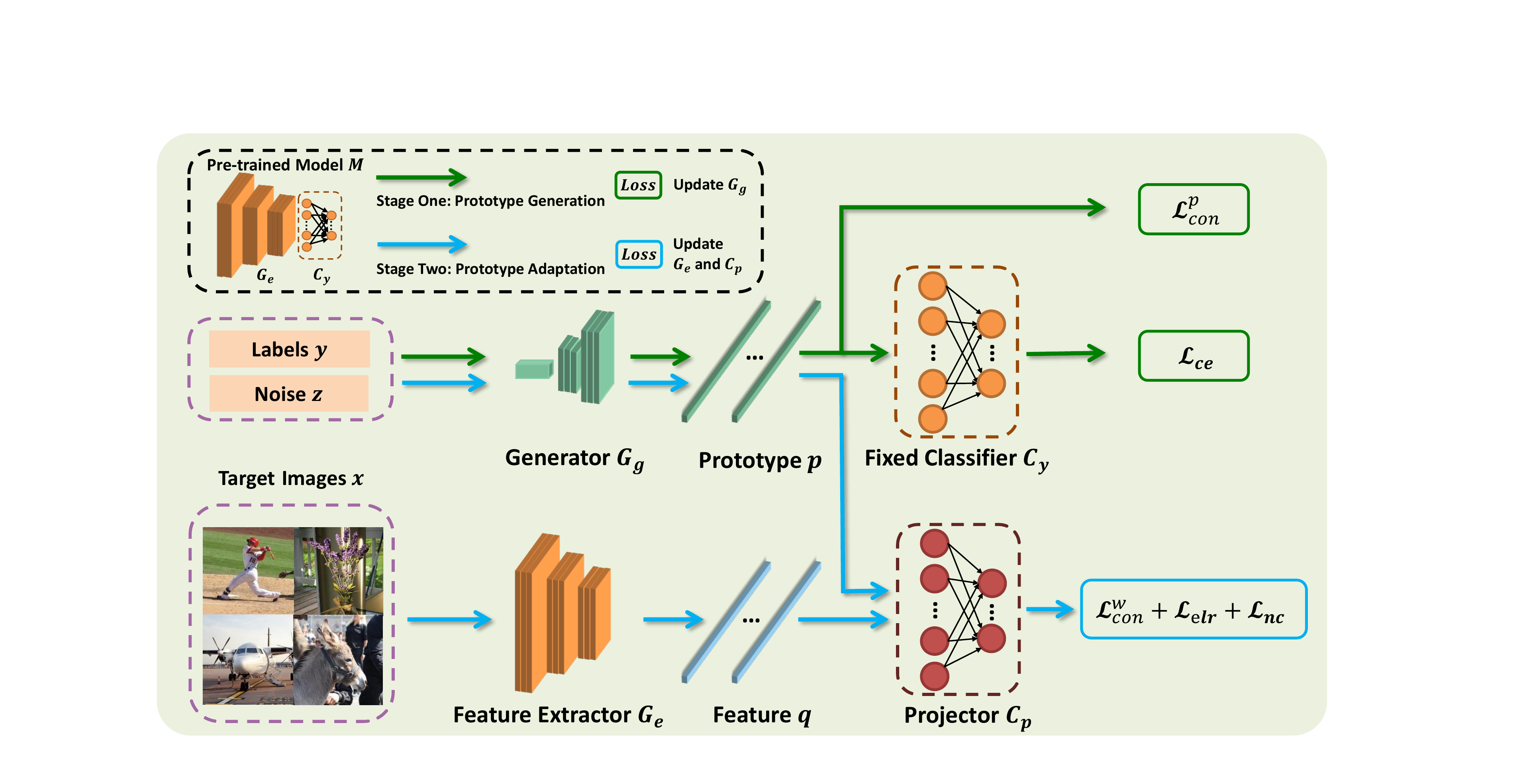}
\vspace{-0.07in}
\caption{An overview of \ournet. \ournet~contains two stages: (1) \textbf{Prototype generation}: under the guidance of the fixed classifier, a generator $G_{g}$ is trained to generate avatar feature prototypes via $\mathcal{L}_{ce}$ and $\mathcal{L}_{con}^{p}$. (2) \textbf{Prototype adaptation}:  in each training batch, we use the learned prototype generator to generate one prototype for each class. Based on the generated prototypes and pseudo labels obtained by clustering, we align each pseudo-labeled target feature to the corresponding class prototype by training a domain-invariant feature extractor via $\mathcal{L}_{con}^{w}$, $\mathcal{L}_{elr}$ and $\mathcal{L}_{nc}$. Note that the classifier $C_{y}$ is fixed during the whole training phase.}
\label{fig:overall}
\vspace{-0.05in}
\end{figure*}

\section{Proposed Method}
\subsection{Problem Definition}
We focus on the task of source-free unsupervised domain adaptation (UDA) in this paper, where only a well-trained source model and unlabeled target data are accessible.
Specifically, we consider a $K$-class classification task, where the source and target domains share with the same label space. We assume that the pre-trained source model consists of a feature extractor $G_{e}$ and a classifier $C_{y}$.  Moreover, we denote the unlabeled target domain by $D_t\small{=}\{\textbf{x}_i\}_{i\small{=}1}^{n_t}$, where $n_{t}$ is the number of target samples.

The key goal is to adapt the source model to the target domain with access to only unlabeled target data. Such a task, however, is very challenging due to 
the lack of source domain data and target domain annotations. Hence, conventional UDA methods requiring source data are unable to tackle this task. To address this task, we innovatively propose a Contrastive Prototype Generation and Adaptation (\ournet)~method.


\subsection{Overall Scheme}
Inspired by that feature prototypes can represent a group of semantically similar instances~\cite{snell2017prototypical}, we explore to generate avatar feature prototypes to represent each source class and use them for class-wise domain alignment. 
As shown in Figure~\ref{fig:overall}, the proposed \ournet~consist of two stages: prototype generation and prototype adaptation.

In the stage one (Section~\ref{stage1}), inspired by that the classifier of the source model contains class distribution information~\cite{Xu2020GenerativeLD}, we propose to train a class conditional generator $G_{g}$ to learn such class information and generate avatar feature prototypes for each class. Meanwhile, we use the source classifier $C_{y}$ to judge whether $G_{g}$ generates correct feature prototypes \wrt classes. By training the generator $G_g$ to confuse $C_{y}$  via both cross-entropy $\mathcal{L}_{ce}$ and the contrastive loss $\mathcal{L}_{con}^{p}$, we are able to generate intra-class compact and inter-class separated feature prototypes.  
Meanwhile, to overcome the lack of target domain annotations, we resort to a self pseudo-labeling strategy to generate pseudo labels for each target data (Section~\ref{sec:pse}).

In the stage two (Section~\ref{stage2}), we adapt the source model to the target by aligning the pseudo-labeled target features to the source prototypes.  Specifically, we conduct class-wise alignment through a contrastive loss $\mathcal{L}_{con}^{w}$ based on a domain projector $C_p$. Meanwhile,  we devise an early learning regularization term $\mathcal{L}_{elr}$ to prevent remembering noisy pseudo labels. Lastly, to make the feature more discriminative, we further impose a neighborhood clustering loss $\mathcal{L}_{nc}$.

The overall training procedure of \ournet~can be summarized as follows:
\begin{equation}
\label{loss:generator}
\min_{\theta_{g}} \mathcal{L}_{ce}(\theta_{g}) + \mathcal{L}_{con}^{p}(\theta_{g}),
\end{equation}
\vspace{-3.5mm}
\begin{equation}
\label{loss:extractor}
\min_{\{\theta_{e}, \theta_{p}\}} \mathcal{L}_{con}^{w}(\theta_{e}, \theta_{p}) + \lambda \mathcal{L}_{elr}(\theta_{e}, \theta_{p}) + \eta \mathcal{L}_{nc}(\theta_{e}),
\end{equation}
where $\theta_{g}$, $\theta_{e}$ and $\theta_{p}$ denotes the parameters of the generator $G_{g}$, the feature extractor $G_{e}$ and the projector $C_{p}$, respectively. Moreover,  $\lambda$ and $\eta$ are  trade-off parameters to balance losses. For simplicity, we set the trade-off parameter to 1  in Eq.~(\ref{loss:generator}) based on our preliminary studies.


\subsection{Contrastive Prototype Generation}\label{stage1}
The absence of the source data makes UDA challenging. To handle this, we propose to generate feature prototypes for each class by exploring the class distribution information hidden in the source classifier~\cite{Xu2020GenerativeLD}. To this end, we use the source classifier $C_{y}$  to train the class conditional generator $G_{g}$. 
To be specific, as shown in Figure~\ref{fig:overall}, given a uniform noise $\textbf{z}\small{\sim} U(0,1)$ and a label $\textbf{y}\small{\in}\mathbb{R}^K$ as inputs, the generator $G_{g}$ first generates the feature prototype $\textbf{p} \small{=} G_{g}(\textbf{y},\textbf{z})$ (More details of the generator and the generation process can be found in Supplementary). Then, the classifier $G_y$ judges whether the generated prototype belongs to $\textbf{y}$ and trains the generator via the cross entropy loss:
\begin{equation}
\label{eq:ce}
\mathcal{L}_{ce} = -\textbf{y}\log C_{y}(\textbf{p}),
\end{equation}
where $\textbf{p}$ is the generated prototype and $C_{y}(\textbf{p})$ denotes the prediction of the classifier.
In this way, the generator is capable of generating feature prototypes for each category.  

\begin{figure}[t]
\centering
\begin{minipage}{0.49\linewidth}
\subfigure[Training with $\mathcal{L}_{ce}$]{
\includegraphics[width=3.5cm,clip,trim={3.6cm 6.5cm 2.5cm 5.5cm}]{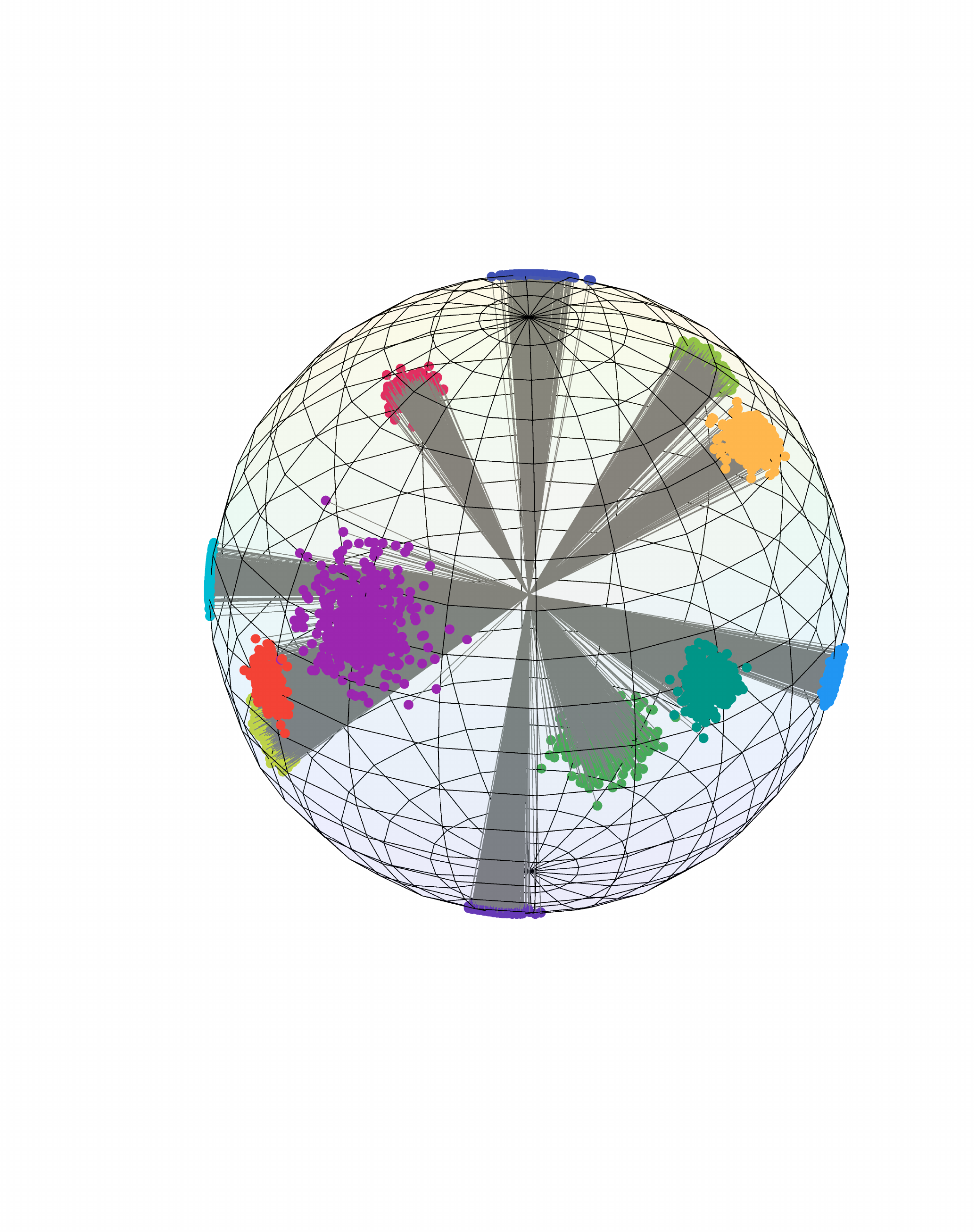}
\label{vis:lce}
}
\end{minipage}
\begin{minipage}{0.49\linewidth}
\subfigure[Training with $\mathcal{L}_{ce}\small{+}\mathcal{L}_{con}^{p}$]{
\includegraphics[width=3.5cm,clip,trim={3.6cm 6.5cm 2.5cm 5.5cm}]{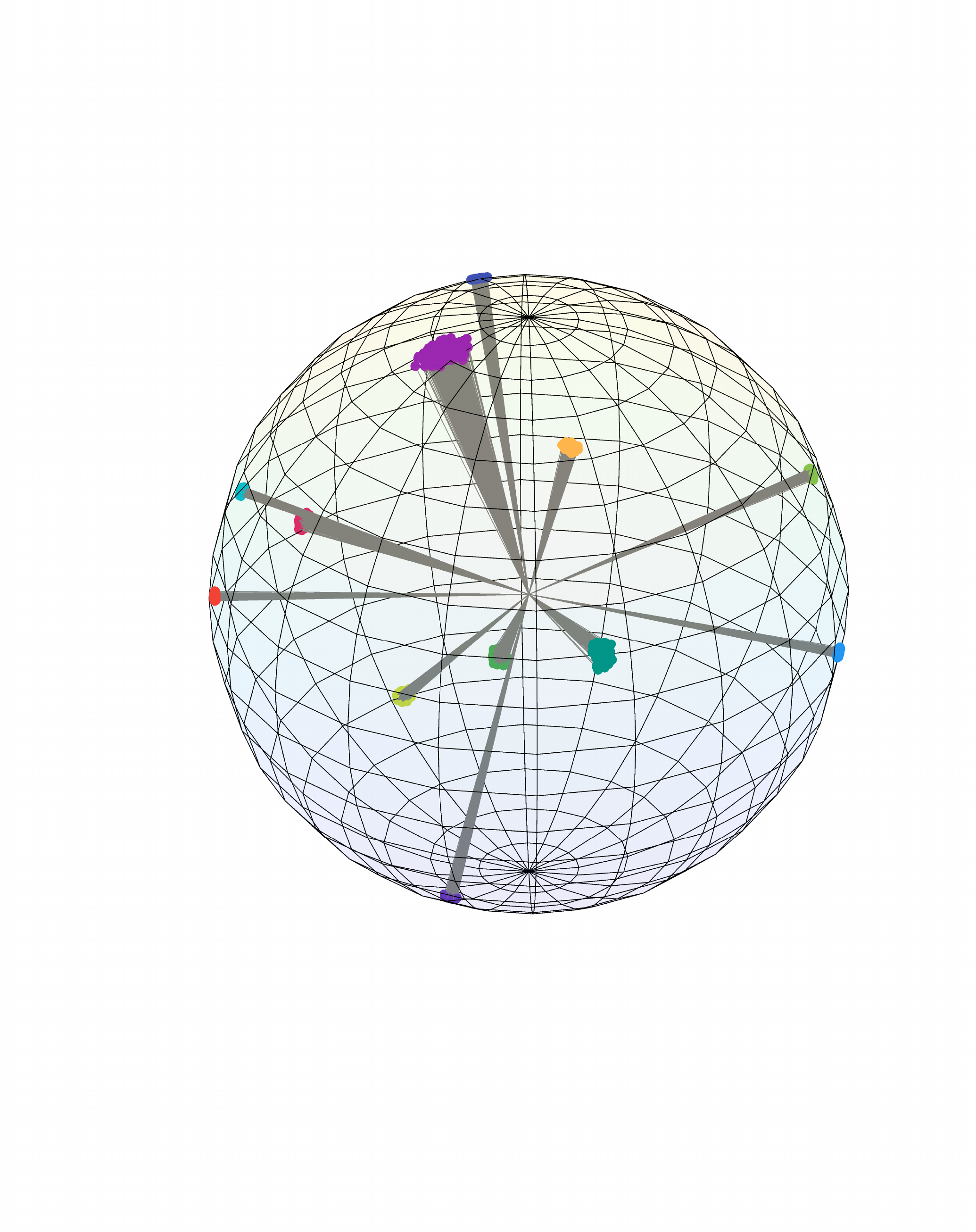}
\label{vis:lcec}
}
\end{minipage}
\label{figdata}
\caption{Visualizations of the generated feature prototypes by the generator  trained with different losses. Compared to training with only the cross entropy   $\mathcal{L}_{ce}$, the contrastive loss $\mathcal{L}_{con}^{p}$
encourages the prototypes of the same category to be more compact and those of different categories to be more separated. Better viewed in color.}
\label{vis:lce_total}
\end{figure}

However, as shown in Figure~\ref{vis:lce}, training the generator with only the cross entropy may make the feature prototypes not well compact and prototypical. As a result, domain alignment with these prototypes may make the adapted model less discriminative, leading to insufficient   performance (See Table~\ref{tab:prototypes}). To address this, motivated by InfoNCE~\cite{Oord2018RepresentationLW,Zhang2021UnleashingTP}, we further impose a contrastive loss for all generated prototypes to encourage more prototypical prototypes:
\begin{equation}
\label{eq:C_p}
\mathcal{L}_{con}^{p}\! \small{=}  \small{-}\log \frac{\exp(\phi(\textbf{p}, \textbf{k}^+)/\tau)}{ \exp(\phi(\textbf{p}, \textbf{k}^+)/\tau) \small{+} \sum_{j=1}^{K\small{-}1}\exp(\phi(\textbf{p}, \textbf{k}_j^-)/\tau)},
\end{equation}
where $\textbf{p}$ denotes any anchor prototype. For each anchor,  we sample  the positive pair  $\textbf{k}^+$ by randomly selecting a generated prototype  with  the same category to the anchor $\textbf{p}$,    and sample $K\small{-}1$ negative pairs $\textbf{k}^-$  that have diverse classes with the anchor. Here,  in each training batch, we generate at least 2 prototypes for each class in the stage one. Moreover, $\phi(\cdot,\cdot)$ denotes the cosine similarity and $\tau$ is a temperature factor.

As shown in Figure~\ref{vis:lcec}, by training the generator with $\mathcal{L}_{ce}\small{+}\mathcal{L}_{con}^{p}$, the generated prototypes are more representative (\ie intra-class compact and   inter-class separated). Interestingly, we empirically observe that the inter-class cosine distance will converge closely to 1 (\ie cosine similarity close to 0) by training with $\mathcal{L}_{ce}\small{+}\mathcal{L}_{con}^{p}$ (See Table~\ref{tab:prototypes}),  if the feature dimensions are larger than the number of classes. That is, the generated prototypes of different categories are approximatively orthometric in the high-dimensional feature space.

\subsection{Pseudo Label Generation for Target Data}
\label{sec:pse}
Domain alignment can be conducted based on the generated avatar source prototypes, However, the alignment is non-trivial due to the lack of target annotations, which makes the class-wise alignment  difficult~\cite{pei2018multi,Kang2019ContrastiveAN}. To address this, we generate pseudo labels based on a self-supervised pseudo-labeling strategy, proposed in~\cite{liang2020shot}. To be specific, let $\textbf{q}_i \small{=}G_{e}(\textbf{x}_i)$ denote  the feature vector and let $\hat{y}_i^k \small{=} C_{y}^{k}(\textbf{q})$ be the predicted probability of the classifier regarding the class $k$. We first attain the initial centroid for each class $k$ by:
\begin{equation}
\textbf{c}_{k} = \frac{\sum_{i=1}^{n_{t}} \hat{y}_i^k  \textbf{q}_i}{\sum_{i=1}^{n_{t}} \hat{y}_i^k},
\end{equation}
where $n_{t}$ is the number of target data. These centroids help to characterize  the distribution of different categories~\cite{liang2020shot}. Then, the pseudo label  of the $i$-th target data is obtained via a nearest centroid approach:
\begin{equation}\label{pseudo}
\bar{y}_{i} = \mathop{\arg\max}_{k}\phi(\textbf{q}_i, \textbf{c}_{k}),
\end{equation}
where $\phi(\cdot,\cdot)$ denotes the cosine similarity, and the pseudo label  $\bar{y}_{i}\small{\in}\mathbb{R}^1$ is a scalar index. 
During the training process, we update the centroid of each class by 
$\textbf{c}_{k}  \small{=} \frac{\sum_{i=1}^{n_{t}} \mathbb{I}(\bar{y}_{i}\small{=}k)   \textbf{q}_i}{\sum_{i=1}^{n_{t}}\mathbb{I}(\bar{y}_{i}\small{=}k)}$ and then update  pseudo labels based on Eqn.~(\ref{pseudo})  in  each   epoch,
where $\mathbb{I}(\cdot)$ is the indicator function.

\subsection{Contrastive Prototype Adaptation}\label{stage2} 
Based on the generated prototypes and target pseudo labels, we conduct prototype adaptation to alleviate domain shifts. Here, in each training batch, we generate one prototype for each class. 
However, due to domain discrepancies, the pseudo labels can be quite noisy, making the adaptation difficult. 
To address this, we propose a new contrastive prototype adaptation strategy, which consists of three key components: (1) weighted contrastive alignment; (2) early learning regularization; (3) target neighborhood clustering. 



\paragraph{Weighted Contrastive Alignment.} 
Based on the pseudo-labeled target data, we then conduct class-wise contrastive learning to align the target data to the corresponding source feature prototype.
However, the  pseudo labels may be noisy, which degrades contrastive alignment. To address this, we propose to differentiate pseudo-labeled target data and assign higher importance to the reliable ones. Motivated by~\cite{Chen2019ProgressiveFA} that reliable samples are generally more close to the class centroid, we compute the confidence weight by:
\begin{equation}
w_{i} = \frac{\exp(\phi(\textbf{q}_i, \textbf{c}_{\bar{y}_{i}})/\tau)}{\sum_{k=1}^{K}\exp(\phi(\textbf{q}_i, \textbf{c}_{k} )/\tau)}, 
\end{equation}  
where the feature with higher similarity to the corresponding centriod will have higher importance. 
Then, we can  conduct weighted contrastive alignment. To this end, inspired by~\cite{chen2020simple}, we first  use a non-linear projector $C_p$ to project the target features and source prototypes  to a $l_2$-normalized contrastive feature space. Specifically, the target contrastive feature is denoted as $\textbf{u} \small{=} C_{p}(\textbf{q})$, while the  prototype contrastive feature is  denoted as  $\textbf{v}  \small{=} C_{p}(\textbf{p})$. Then, for any target feature $\textbf{u}_{i}$ as an anchor, we conduct prototype adaptation via a weighted contrastive loss:
\begin{equation}
\label{eqn_lwc}
\mathcal{L}_{con}^{w} \!=\! -\! w_{i}\!\log\frac{\exp(\textbf{u}_{i}^{\top} \textbf{v}^+/\tau )}{\exp(\textbf{u}_{i}^{\top}  \textbf{v}^+/\tau) \small{+} \sum_{j=1}^{K\small{-}1}\exp(\textbf{u}_{i}^{\top} \textbf{v}^-_{j}/\tau)},
\end{equation}
where the positive pair  $\textbf{v}^+$ is the prototype with the same class to the anchor $\textbf{u}_{i}$, while the negative pairs $\textbf{v}^-$ are the prototypes with different classes. 

\paragraph{Early Learning Regularization.}
\label{sec:lreg} 
To further prevent the model from memorizing noise, we propose to regularize the learning process via an early learning regularizer. Since DNNs first memorize the clean samples with correct labels and then the noisy data with wrong labels~\cite{arpit2017closer}, the model in the “early learning” phase can be more predictable to the noisy data. Therefore, we seek to use the early predictions of each sample to regularize learning.
To this end, we devise a memory bank $\mathcal{H}\small{=}\{\textbf{h}_{1}, \textbf{h}_{2},...,\textbf{h}_{n_t}\}$ to record non-parametric predictions of each target sample, and update them based on new predictions via a momentum strategy. Formally, for the $i$-th sample, we predict its non-parametric prediction regarding the $k$-th prototype by  
$o_{i,k} \small{=} \frac{\exp(\textbf{u}_{i}^{\top}\textbf{v}_k/\tau)}{\sum_{j=1}^{K}\exp(\textbf{u}_{i}^{\top} \textbf{v}_{j}/\tau)}$,
 and update the momentum by:
\begin{equation}
\textbf{h}_{i} \xleftarrow{} \beta \textbf{h}_{i} + (1-\beta)\textbf{o}_{i}, 
\end{equation}
where $\textbf{o}_{i}\small{=}[o_{i,1},...,o_{i,K}]$, and $\beta$ denotes the momentum coefficient. Based on the memory bank, for the $i$-th data, we further train the model via  an early learning regularizer $\mathcal{L}_{elr}$, proposed in~\cite{liu2020early}:
\begin{equation}
\label{loss:reg}
\mathcal{L}_{elr} = \log(1-\textbf{o}_{i}^{\top} \textbf{h}_{i}).
\end{equation}
This regularizer enforces the current prediction to be close to the prediction momentum, which helps to prevent overfitting to label noise. Note that the use of $\mathcal{L}_{elr}$ in this paper is different from~\cite{liu2020early}, which focuses on classification tasks and uses parametric predictions.

\paragraph{Target Neighborhood Clustering.} 
To enhance the contrastive alignment, we further resort to feature clustering to make the target features more compact. Inspired by~\cite{Saito2020UniversalDA} that the intra-class samples in the same domain are generally more close, we propose to close the distance between each target sample and its nearby neighbors.  
To this end, we maintain a memory bank $\mathcal{Q}\small{=} \{\textbf{q}_{1}, \textbf{q}_{2}, ...,\textbf{q}_{n_{t}}\}$ to restore all target features, which are updated when new features are extracted  in each iteration. 
Based on the bank, for the $i$-th sample's feature $\textbf{q}_i$, we can compute its normalized similarity with any feature $\textbf{q}_j$ by 
$\textbf{s}_{i,j} \small{=} \frac{\exp(\phi(\textbf{q}_i,\textbf{q}_j)/\tau)}{\sum_{l=1, l\neq i}^{n_{t}}\exp(\phi(\textbf{q}_i, \textbf{q}_{l})/\tau)}$.
Motivated by that minimizing the entropy of the normalized similarity helps to learn compact features for similar data~\cite{Saito2020UniversalDA},  we further train the extractor via a neighborhood clustering loss:
\begin{equation}
\label{loss:nc}
\mathcal{L}_{nc} = -\sum_{j=1, j\neq i}^{n_t} \textbf{s}_{i, j} \log(\textbf{s}_{i, j}).
\end{equation}
Note that the entropy minimization here does not use pseudo labels, so the learned compact target features are (to some degree) robust to pseudo label noise. We summarize the overall training scheme of \ournet~in Algorithms~\ref{al:training}, while the inference is provided in the supplementary.

\begin{algorithm}[t]
    \small
    \caption{Training of \ournet}\label{al:training}
    \begin{algorithmic}[1]
        \REQUIRE Unlabeled target data $D_t\small{=}\{\textbf{x}_i\}_{i=1}^{n_t}$; Source model $\{G_{e}, C_{y}\}$; Training epoch $E$, $M$; Parameters $\eta$, $\beta$, $\tau$, $\lambda$.
    \ENSURE Projector $C_{p}$; Generator $G_g$. 
    \FOR{$e = 1 \to E$}
        \STATE Generate prototypes $\textbf{p}$ based on $G_{g}$;
        \STATE Compute $\mathcal{L}_{ce}$ and $\mathcal{L}_{con}^{p}$ based on Eqns.~(\ref{eq:ce}) and (\ref{eq:C_p});
        \STATE loss.backward()  based on Eqn.~(\ref{loss:generator}).
    \ENDFOR
    \FOR{$m = 1 \to M$}
        \STATE Generate prototypes $\textbf{p}$ for each class based on fixed $G_{g}$;
        \STATE Extract target data features $G_{e}(\textbf{x})$ based on $G_e$;
        \STATE Obtain target pseudo labels based on Eqn.~(\ref{pseudo});
        \STATE Obtain contrastive features $\textbf{h}_{t}$ based on $C_{p}$; 
        \STATE Compute $\mathcal{L}_{con}^{w}$, $\mathcal{L}_{elr}$, $\mathcal{L}_{nc}$ based on Eqns.~(\ref{eqn_lwc}), (\ref{loss:reg}), (\ref{loss:nc});
        \STATE loss.backward()  based on Eqn.~(\ref{loss:extractor}).
    \ENDFOR
    \RETURN $G_{e}$ and $C_{y}$.
     \end{algorithmic}
\end{algorithm}


\section{Experiments} 




\paragraph{Datasets.}\label{des:dataset}

We conduct the experiments on three benchmark datasets:
(1) \textbf{Office-31}~\cite{Saenko2010AdaptingVC} is a standard domain adaptation dataset that is made up of three distinct domains, \ie Amazon (A), Webcam (W) and DSLR (D). Three domains share 31 categories and contain 2817, 795 and 498 samples, respectively.
(2) \textbf{VisDA}~\cite{Peng2017VisDATV} is a large-scale challenging dataset that concentrates on the 12-class synthesis-to-real object recognition task. The source domain contains 152k synthetic images
while the target domain has 55k real object images. 
(3) \textbf{Office-Home}~\cite{Venkateswara2017DeepHN} is a medium-sized dataset, which contains four distinct domains, \ie Artistic images (Ar), Clip Art (Cl), Product images (Pr) and Real-world images (Rw). Each of the four domains has 65 categories.

\paragraph{Baselines.}

We compare \ournet~with three types of baselines:
(1) source-only: ResNet~\cite{He2016DeepRL};
(2) unsupervised domain adaptation with source data:
 MCD~\cite{saito2018maximum}, CDAN~\cite{long2018conditional}, TPN~\cite{pan2019transferrable}, SAFN~\cite{xu2019larger}, SWD~\cite{lee2019sliced}, MDD~\cite{zhang2019bridging}, CAN~\cite{Kang2019ContrastiveAN}, DMRL~\cite{wu2020dual}, BDG~\cite{yang2020bi}, PAL~\cite{hu2020panda}, MCC~\cite{jin2020minimum}, SRDC~\cite{tang2020unsupervised}; (3) source-free unsupervised domain adaptation: SHOT~\cite{liang2020shot}, PrDA~\cite{kim2020progressive}, MA~\cite{Li2020ModelAU} and BAIT~\cite{Yang2020UnsupervisedDA}.

\paragraph{Implementation Details.}

We implement our method based on PyTorch\footnote{The source code is available:~\url{github.com/SCUT-AILab/CPGA}.}. For a fair comparison, we report the results of all baselines in the corresponding papers.
For the network architecture, we adopt a ResNet~\cite{He2016DeepRL}, pre-trained on ImageNet, as the backbone of all methods. Following~\cite{liang2020shot}, we replace the original fully connected (FC) layer with a task-specific FC layer followed by a weight normalization layer. 
The projector consists of three FC layers with hidden feature dimensions of 1024, 512 and 256.
We train the source model via label smoothing technique~\cite{Mller2019WhenDL} and  train \ournet~using SGD optimizer.
We set the learning rate and epoch to 0.01 and 40 for VisDA and to 0.001 and 400 for Office-31 and Office-Home.
For hyper-parameters, we set $\eta$, $\beta$, $\tau$ and batch size to 0.05, 0.9, 0.07 and 64, respectively. Besides, we set $\lambda \small{=}7$ for Office-31 and Office-home while $\lambda \small{=} 5$ for VisDA. Following~\cite{Xu2020GenerativeLD}, the dimension of noise $\textbf{z}$ is 100.
We put more implementation details in the supplementary.

\begin{table}[t]
\setlength\tabcolsep{1pt}
    \begin{center}
    \scalebox{0.7}{
         \begin{tabular}{lcccccccl}
         \toprule
         Method & Source-free & A$\rightarrow$D & A$\rightarrow$W & D$\rightarrow$W & W$\rightarrow$D & D$\rightarrow$A & W$\rightarrow$A & Avg.\\
         \midrule
         ResNet-50~\cite{He2016DeepRL} & \xmark & 68.9 & 68.4 & 96.7 & 99.3 & 62.5 & 60.7 & 76.1 \\
         MCD~\cite{saito2018maximum} & \xmark & 92.2 & 88.6 & 98.5 & 100.0 & 69.5 & 69.7 & 86.5 \\
         CDAN~\cite{long2018conditional} & \xmark & 92.9 & 94.1 & 98.6 & 100.0 & 71.0 & 69.3 & 87.7 \\
         MDD~\cite{zhang2019bridging} & \xmark & 90.4 & 90.4 & 98.7 & 99.9 & 75.0 & 73.7 & 88.0 \\
         CAN~\cite{Kang2019ContrastiveAN} & \xmark & 95.0 & 94.5 & 99.1 & 99.6 & 70.3 & 66.4 & 90.6 \\
         DMRL~\cite{wu2020dual} & \xmark & 93.4 & 90.8 & 99.0 & 100.0 & 73.0 & 71.2 & 87.9 \\
         BDG~\cite{yang2020bi} & \xmark & 93.6 & 93.6 & 99.0 & 100.0 & 73.2 & 72.0 & 88.5 \\
         MCC~\cite{jin2020minimum} & \xmark & 95.6 & 95.4 & 98.6 & 100.0 & 72.6 & 73.9 & 89.4 \\
         SRDC~\cite{tang2020unsupervised} & \xmark & 95.8 & 95.7 & 99.2 & 100.0 & 76.7 & 77.1 & 90.8 \\
         \midrule
         PrDA~\cite{kim2020progressive} & \cmark & 92.2 & 91.1 & 98.2 & 99.5 & 71.0 & 71.2 & 87.2 \\
         SHOT~\cite{liang2020shot} & \cmark & 93.1 & 90.9 & \textbf{98.8} & 99.9 & 74.5 & 74.8 & 88.7 \\
         BAIT~\cite{Yang2020UnsupervisedDA} & \cmark & 92.0 & \textbf{94.6} & 98.1 & \textbf{100.0} & 74.6 & 75.2 & 89.1 \\
         MA~\cite{Li2020ModelAU} & \cmark & 92.7 & 93.7 & 98.5 & 99.8 & 75.3 & \textbf{77.8} & 89.6 \\
         \midrule
         \ournet~(ours) & \cmark & \textbf{94.4} & 94.1 & 98.4 & 99.8 & \textbf{76.0} & 76.6 & \textbf{89.9} \\
         \bottomrule
         \end{tabular}
         }
    \end{center}
    \vspace{-0.05in}
    \caption{\label{tab:office}Accuracy (\%) on the small-sized \textbf{Office-31} (ResNet-50).}
    \vspace{-0.1in}
\end{table}

\begin{table*}[!hbt]
\setlength\tabcolsep{3pt}
    \begin{center}
    \scalebox{0.75}{
         \begin{tabular}{lcccccccccccccc}
         \toprule
         Method & Source-free & plane & bicycle & bus & car & horse & knife & mcycl & person & plant & sktbrd & train & truck & Per-class\\
         \midrule
         ResNet-101~\cite{He2016DeepRL} & \xmark & 55.1 & 53.3 & 61.9 & 59.1 & 80.6 & 17.9 & 79.7 & 31.2 & 81.0 & 26.5 & 73.5 & 8.5 & 52.4 \\
         CDAN~\cite{long2018conditional} & \xmark & 85.2 & 66.9 & 83.0 & 50.8 & 84.2 & 74.9 & 88.1 & 74.5 & 83.4 & 76.0 & 81.9 & 38.0 & 73.9 \\
         SAFN~\cite{xu2019larger} & \xmark & 93.6 & 61.3 & 84.1 & 70.6 & 94.1 & 79.0 & 91.8 & 79.6 & 89.9 & 55.6 & 89.0 & 24.4 & 76.1 \\
         SWD~\cite{lee2019sliced} & \xmark & 90.8 & 82.5 & 81.7 & 70.5 & 91.7 & 69.5 & 86.3 & 77.5 & 87.4 & 63.6 & 85.6 & 29.2 & 76.4 \\
         TPN~\cite{pan2019transferrable} & \xmark & 93.7 & 85.1 & 69.2 & 81.6 & 93.5 & 61.9 & 89.3 & 81.4 & 93.5 & 81.6 & 84.5 & 49.9 & 80.4 \\
         PAL~\cite{hu2020panda} & \xmark & 90.9 & 50.5 & 72.3 & 82.7 & 88.3 & 88.3 & 90.3 & 79.8 & 89.7 & 79.2 & 88.1 & 39.4 & 78.3 \\    
         MCC~\cite{jin2020minimum} & \xmark & 88.7 & 80.3 & 80.5 & 71.5 & 90.1 & 93.2 & 85.0 & 71.6 & 89.4 & 73.8 & 85.0 & 36.9 & 78.8 \\
         CoSCA~\cite{Dai_2020_ACCV} & \xmark & 95.7 & 87.4 & 85.7 & 73.5 & 95.3 & 72.8 & 91.5 & 84.8 & 94.6 & 87.9 & 87.9 & 36.8 & 82.9 \\
         \midrule
         PrDA~\cite{kim2020progressive} & \cmark & 86.9 & 81.7 & \textbf{84.6} & 63.9 & \textbf{93.1} & 91.4 & 86.6 & 71.9 & 84.5 & 58.2 & 74.5 & 42.7 & 76.7\\
         SHOT~\cite{liang2020shot} & \cmark & 92.6 & 81.1 & 80.1 & 58.5 & 89.7 & 86.1 & 81.5 & 77.8 & 89.5 & 84.9 & 84.3 & 49.3 & 79.6\\
         MA~\cite{Li2020ModelAU} & \cmark & 94.8 & 73.4 & 68.8 & \textbf{74.8} & \textbf{93.1} & 95.4 & 88.6 & \textbf{84.7} & 89.1 & 84.7 & 83.5 & 48.1 & 81.6\\
         BAIT~\cite{Yang2020UnsupervisedDA} & \cmark & 93.7 & 83.2 & 84.5 & 65.0 & 92.9 & 95.4 & 88.1 & 80.8 & 90.0 & 89.0 & 84.0 & 45.3 & 82.7 \\
        \midrule
        \ournet~(ours, 40 epochs) & \cmark & 94.8 & 83.6 & 79.7 & 65.1 & 92.5 & 94.7 & \textbf{90.1} & 82.4 & 88.8 & 88.0 & \textbf{88.9} & {60.1} & {84.1} \\
        \ournet~(ours, 400 epochs) & \cmark & \textbf{95.6} & \textbf{89.0} & 75.4 & 64.9 & 91.7 & \textbf{97.5} & 89.7 & 83.8 & \textbf{93.9} & \textbf{93.4} & 87.7 & \textbf{69.0} & \textbf{86.0} \\
        \bottomrule
        \end{tabular}
    }
    \end{center}
    \vspace{-0.1in}
    \caption{\label{tab:visda}Classification accuracies (\%) on the large-scale \textbf{VisDA} dataset (ResNet-101).}
\end{table*}

\subsection{Comparison with State-of-the-arts}

In this section, we compare our proposed \ournet~with the state-of-the-art methods.
For \textbf{Office-31}, as shown in Table~\ref{tab:office}, the proposed \ournet~achieves the best performance compared with source-free UDA methods \wrt the average accuracy over 6 transfer tasks. Moreover, our method shows its superiority in the task of A$\rightarrow$D and D$\rightarrow$A and comparable results on the other tasks.
Note that even compared with the state-of-the-art methods using source data (\eg SRDC), our \ournet~is able to obtain a competitive result as well.
Besides, from Table~\ref{tab:visda}, \ournet~outperforms all the state-of-the-art methods \wrt the average accuracy (\ie per-class accuracy) on the more challenging dataset \textbf{VisDA}. Specifically, \ournet~gets the best accuracy in the \mata{eight} categories and obtains comparable results in others.
Moreover, our \ournet~is able to surpass the baseline methods with source data (\eg CoSCA), which demonstrates the superiority of our proposed method.
For \textbf{Office-Home}, we put the results in the supplementary.

\subsection{Ablation Study}

To evaluate the effectiveness of the proposed two modules (\ie prototype generation and prototype adaptation) and the sensitivity of hyper-parameters, we conduct a series of ablation studies on VisDA.

\paragraph{Effectiveness of Prototype Generation.}

\begin{table}[t]
\setlength\tabcolsep{1pt}
    \begin{center}
    \scalebox{0.8}{
         \begin{tabular}{lccccccc}
         \toprule
         Method & A$\rightarrow$D & A$\rightarrow$W & D$\rightarrow$W & W$\rightarrow$D & D$\rightarrow$A & W$\rightarrow$A & Avg.\\
         \midrule
         DANN (with source data) & 79.7 & 82.0 & 96.9 & 99.1 & 68.2 & 67.4 & 82.2 \\
         DANN (with prototypes) & \textbf{83.7} & 81.1 & \textbf{97.5} & \textbf{99.8} & 63.4 & 63.6 & 81.5 \\
         \midrule
         DMAN (with source data) & 83.3 & 85.7 & 97.1 & 100.0 & 65.1 & 64.4 & 82.6 \\
         DMAN (with prototypes) & \textbf{86.3} & 84.2 & \textbf{97.7} & \textbf{100.0} & 64.7 & \textbf{64.5} & \textbf{82.9} \\
         \midrule
         ADDA (with source data) & 82.9 & 79.9 & 97.4 & 99.4 & 64.9 & 63.6 & 81.4 \\
         ADDA (with prototypes) & \textbf{83.5} & \textbf{81.9} & 97.2 & \textbf{100.0} & 63.8 & 63.0 & \textbf{81.6} \\
         \bottomrule
         \end{tabular}
         }
    \end{center}
    \vspace{-0.1in}
    \caption{\label{tab:proto-office}Comparisons of the existing domain adaptation methods with source data or prototypes on \textbf{Office-31} (ResNet-50).}
\end{table}

In this section, we verify the effect of our generated prototypes in the existing domain adaptation methods (\eg DANN~\cite{ganin2015unsupervised}, ADDA~\cite{Tzeng2017AdversarialDD} and DMAN~\cite{zhang2019whole}), which, previously, cannot solve the domain adaptation problem without source data.
To this end, we introduce our prototype generation module to replace their source data-oriented parts. From Table~\ref{tab:proto-office},
based on prototypes, the existing methods achieve competitive performance compared with the counterparts using source data, or even perform better in some tasks. It demonstrates the superiority and applicability of our prototype generation scheme.

\paragraph{Ablation Studies on Prototype Generation.}

\begin{table}[t]
\vspace{-0.01in}
\renewcommand\arraystretch{1.4}
\setlength\tabcolsep{4pt}
    \begin{center}
         \scalebox{0.8}{\begin{tabular}{cccc}
         \toprule
         Objective & Inter-class distance & Intra-class distance & Per-class (\%)\\
         \midrule
         $\mathcal{L}_{ce}$  & 0.7860 & $3.343\times e^{-4}$ & 85.0\\
         $\mathcal{L}_{ce} + \mathcal{L}_{con}^{p}$ & 1.0034 & $2.670\times e^{-6}$ & 86.0\\
         \bottomrule
         \end{tabular}
         }
    \end{center}
    \vspace{-0.1in}
    \caption{\label{tab:prototypes}Ablation studies on prototype generation in the stage one with different losses. Inter-class distance and intra-class distance is based on cosine distance (range from 0 to 2). \mata{We report per-class accuracy (\%) after training the model on \textbf{VisDA} for 400 epochs}.}
\end{table}

To study the impact of our contrastive loss $\mathcal{L}_{con}^p$, we compare the generated prototype results from models with and without $\mathcal{L}_{con}^p$.
From Table~\ref{tab:prototypes}\footnote{Figure~\ref{vis:lce_total} shows the corresponding visual results of Table~\ref{tab:prototypes}.}, compared with training by cross-entropy loss $\mathcal{L}_{ce}$ only, optimizing the generator via $\mathcal{L}_{ce} \small{+} \mathcal{L}_{con}^{p}$ makes the inter-class features separated (\ie larger inter-class distance) and intra-class features compact (\ie smaller intra-class distance).
The $\mathcal{L}_{con}^p$ loss also helps to enhance the performance from $85.0\%$ to $86.0\%$.


\paragraph{Ablation Studies on Prototype Adaptation.}

\begin{table}[t]
    \begin{center}
    \scalebox{0.8}{
         \begin{tabular}{ccccc|c}
         \toprule
         Backbone & $\mathcal{L}_{con}$& $\mathcal{L}_{con}^{w}$ & $\mathcal{L}_{elr}$ & $\mathcal{L}_{nc}$ & Per-class (\%) \\
         \midrule
         \cmark &  &  &  &  & 52.4 \\
         \cmark & \cmark &  &  &  & 80.9 \\
         \cmark &  & \cmark &  &  &  82.7\\
         \cmark &  & \cmark & \cmark &  & 85.4 \\
         \cmark &  & \cmark & \cmark & \cmark & \textbf{86.0} \\
         \bottomrule
         \end{tabular}
         }
    \end{center}
    \vspace{-0.1in}
    \caption{\label{tab:ablation}Ablation study for the losses (\ie $\mathcal{L}_{con}^w$, $\mathcal{L}_{elr}$ and $\mathcal{L}_{nc}$) of prototype adaptation. \mata{We show the per-class accuracy (\%) of the model trained on \textbf{VisDA} for 400 epochs.} $\mathcal{L}_{con}$ denotes $\mathcal{L}_{con}^{w}$ without confidence weight $w$.}
\end{table}

To investigate the losses of prototype adaptation, we show the quantitative results of the models optimized by different losses.
As shown in Table~\ref{tab:ablation}, compared with the conventional contrastive loss $\mathcal{L}_{con}$, our proposed contrastive loss $\mathcal{L}_{con}^w$ achieves a more promising result on VisDA. \mata{Such a result verifies the ability of alleviating pseudo label noise of the confidence weight $w$.}
Besides, our model has the ability to further improve the performance when introducing the losses $\mathcal{L}_{elr}$ and $\mathcal{L}_{nc}$.
When combining all the three losses (\ie $\mathcal{L}_{con}^w$, $\mathcal{L}_{elr}$ and $\mathcal{L}_{nc}$), we obtain the best performance.

\begin{table}[t]
\vspace{-0.05in}
\renewcommand\arraystretch{1.00}
\setlength\tabcolsep{2.1pt}
\begin{center}
  \scalebox{0.8}{
    \begin{tabular}{ccccccccccc}
    \toprule
    \multirow{2}{*}{Parameter} &
    \multicolumn{5}{c}{$\lambda$}&
    \multicolumn{5}{c}{$\eta$}\cr
    \cmidrule(lr){2-6} \cmidrule(lr){7-11}
    &1 & 3& 5& 7& 9&
    0.001& 0.005& 0.01 &0.05&0.1\cr
    \midrule
    Acc. (40 epochs) & 83.2 & 83.9 & \textbf{84.1} & 83.3 & 82.2 & 82.7& 83.1 & 83.3 & \textbf{84.1} &81.0 \cr
    Acc. (400 epochs) & 83.3 & 85.0 & \textbf{86.0} & 85.5 & 85.3 & 85.5 & 85.6 & 85.5 & \textbf{86.0} &83.0 \cr
    \bottomrule
    \end{tabular}
    }
    \end{center}
    \vspace{-0.1in}
    \caption{
 \label{tab:para_sen}Influence of the trade-off parameter $\lambda$ and $\eta$ in terms of per-class accuracy (\%) on \textbf{VisDA}. The value of $\lambda$ is chosen from $[1, 3, 5, 7, 9]$ and $\eta$ is chosen from $[0.001, 0.005, 0.01, 0.05, 0.1]$. In each experiment, the rest of hyper-parameters are fixed.
  }
\end{table}

\paragraph{Influence of Hyper-parameters.}

In this section, we evaluate the sensitivity of two hyper-parameters $\lambda$ and $\eta$ on VisDA via an unsupervised reverse validation strategy~\cite{ganin2016domain} based on the source prototypes. 
For convenience, we set $\eta=0.05$ when studying $\lambda$, and set $\lambda=5$ when studying $\eta$.
As shown in Table~\ref{tab:para_sen}, the proposed method achieves the best performance when setting $\lambda = 5$ and $\eta = 0.05$ on VisDA. 
The results also demonstrate that our method is non-sensitive for the hyper-parameters.
Besides, we put more analysis of hyper-parameters in the supplementary.

\section{Conclusions}
This paper has proposed a prototype generation and adaptation  (namely \ournet) method for source-free UDA. Specifically, we overcome the  lack of source data by generating avatar  feature prototypes for each class via contrastive learning. Based on the generated prototypes, we develop a robust contrastive prototype adaptation strategy to pull the pseudo-labeled target data toward the corresponding source prototypes. In this way, \ournet~adapts the source model to the target domain without access to any source data. Extensive experiments verify the effectiveness and superiority of \ournet.

\section*{Acknowledgments}
This work was partially supported by Key Realm R\&D Program of Guangzhou (202007030007), National Natural Science Foundation of China
(NSFC) 62072190, Program for Guangdong Introducing Innovative and Enterpreneurial Teams 2017ZT07X183, Fundamental Research Funds for the Central Universities D2191240, Guangdong Natural Science Foundation Doctoral Research Project (2018A030310365), International Cooperation Open Project of State Key Laboratory of Subtropical Building Science, South China University of Technology (2019ZA02).

\balance
\bibliographystyle{named}
\bibliography{ijcai21}

\newpage

\section*{Appendix}
In this appendix, we provide the algorithm of inference scheme (Section~\ref{sec:inference}), \mata{more} implementation details (Section~\ref{sec:implement}), and more experimental results (Section~\ref{sec:exp}).

\section*{A. Inference Details of \ournet} \label{sec:inference}
In this section, we present the pseudo-code of \ournet~during inference. Specifically, when getting a well-trained \ournet, we can obtain the target prediction based on the feature extractor $G_{e}$ and the classifier $C_{y}$.
As shown in Algorithm~\ref{al:inference}, given an input image $\mathbf{x}$, we first capture the corresponding feature $G_{e}(\textbf{x})$ and then feed the feature into the classifier $C_{y}$ to generate the target prediction.

\begin{algorithm}
    \small
    \caption{Inference of \ournet}\label{al:inference}
    \begin{algorithmic}[1]
    \REQUIRE Target data $\textbf{x}$, feature extractor $G_{e}$ and classifier $C_{y}$.
    \STATE Extract feature $G_{e}(\textbf{x})$ regrading $\textbf{x}$ using $G_{e}$;
    \STATE Compute the prediction $C_{y}(G_{e}(\textbf{x}))$ using $C_{y}$;
    \RETURN $C_{y}(G_{e}(\textbf{x}))$.
    \end{algorithmic}
\end{algorithm}

\section*{B. More Implementation Details} \label{sec:implement}

\paragraph{Generator Architecture.}

As shown in Table~\ref{tab:generator}, the generator consists of an embedding layer, two FC layers and two deconvolution layers. 
Similar to ACGAN~\cite{odena2017conditional}, given an input noise $\textbf{z}\small{\sim} U(0,1)$ and a label $\textbf{y}\small{\in}\mathbb{R}^K$, we first map the label into a vector using the embedding layer. After that, we combine the vector with the given noise by a element-wise multiplication and then feed it into the following layers.
Since we propose to obtain feature prototypes instead of images, we reshape the output of the generator into a feature vector with the same dimensions as the last FC layer.

\paragraph{Training.}

In the stage one, we train the generator by optimizing $\mathcal{L}_{ce}\small{+}\mathcal{L}_{con}^{p}$.
The batchsize is set to 128.
We use the SGD optimizer with learning rate $=$ 0.001.
In the stage two, to achieve class-wise domain alignment, we generate feature prototypes for K classes in each epoch.

\paragraph{Optional Hyper-parameter Selection.} 
Following~\cite{ganin2016domain}, we select the hyper-parameters via an unsupervised reverse validation strategy. Such a strategy consists of two steps: (1) We generate source prototypes for K classes and predicted labels for the target domain via a well-trained \ournet. (2) We train another \ournet~with pseudo-labeled target data served as the source domain and evaluate the model on the source prototypes. By the end, we obtain the corresponding hyper-parameters based on the best accuracy on source prototypes.

\begin{table*}[t]
\renewcommand\arraystretch{1.5}
    \begin{center}
    \scalebox{0.8}{
        \begin{tabular}{c|c|c|c|c|c|c}
        \hline
        \hline
            \specialrule{0em}{1pt}{1pt}
            \multicolumn{7}{c}{ Backbone Network } \\ 
        \hline
        \hline
        Part & \multicolumn{2}{c|}{ Input $\rightarrow$ Output } & Kernel & Padding & Stride & Activation \\
        \hline
            Embedding & \multicolumn{2}{c|}{ $(batch\_size, 1)$ $\rightarrow$ $(batch\_size, 100)$ } & - & - & - & -\\
            \hline
            Linear & \multicolumn{2}{c|}{ $(batch\_size, 100)$ $\rightarrow$ $(batch\_size, 1024)$ } & - & - & - & ReLU\\
            \hline
            BatchNorm1d & \multicolumn{2}{c|}{ $(batch\_size, 1024)$ $\rightarrow$ $(batch\_size, 1024)$ } & - & - & - & -\\
            \hline
            Linear & \multicolumn{2}{c|}{ $(batch\_size, 1024)$ $\rightarrow$ $(batch\_size, \frac{d}{4}*7*7)$ } & - & - & - & ReLU\\
            \hline
            BatchNorm1d & \multicolumn{2}{c|}{ $(batch\_size, \frac{d}{4}*7*7)$ $\rightarrow$ $(batch\_size, \frac{d}{4}*7*7)$ } & - & - & - & -\\
            \hline
            Reshape & \multicolumn{2}{c|}{ $(batch\_size, \frac{d}{4}*7*7)$ $\rightarrow$ $(batch\_size, \frac{d}{4}, 7, 7)$ } & - & - & - & -\\
            \hline
            ConvTranspose2d & \multicolumn{2}{c|}{ $(batch\_size, \frac{d}{4}, 7, 7)$ $\rightarrow$ $(batch\_size, \frac{d}{8}, 6, 6)$ } & 2 & 1 & 2 & -\\
            \hline
            BatchNorm2d & \multicolumn{2}{c|}{ $(batch\_size, \frac{d}{8}, 6, 6)$ $\rightarrow$ $(batch\_size, \frac{d}{8}, 6, 6)$ } & - & - & - & ReLU\\
            \hline
            ConvTranspose2d & \multicolumn{2}{c|}{ $(batch\_size, \frac{d}{8}, 6, 6)$ $\rightarrow$ $(batch\_size, \frac{d}{16}, 4, 4)$ } & 3 & 1 & 2 & -\\
            \hline
            BatchNorm2d & \multicolumn{2}{c|}{ $(batch\_size, \frac{d}{16}, 4, 4)$ $\rightarrow$ $(batch\_size, \frac{d}{16}, 4, 4)$ } & - & - & - & ReLU\\
            \hline
            Reshape & \multicolumn{2}{c|}{ $(batch\_size, \frac{d}{16}, 4, 4)$ $\rightarrow$ $(batch\_size, d)$ } & - & - & - & -\\
            \hline
            \hline
         \end{tabular}}
    \end{center}
    \caption{\label{tab:generator}Detailed architecture of the generator, where $d$ denote the output dimensions, \eg 2048.}
\end{table*}

\begin{table*}[!h]
\vspace{0.4cm}
\setlength\tabcolsep{3pt}
    \begin{center}
    \scalebox{0.67}{  
         \begin{tabular}{lcccccccccccccc}
         \toprule
         Method & Source-free & Ar$\rightarrow$Cl & Ar$\rightarrow$Pr & Ar$\rightarrow$Rw & Cl$\rightarrow$Ar & Cl$\rightarrow$Pr & Cl$\rightarrow$Rw & Pr$\rightarrow$Ar & Pr$\rightarrow$Cl & Pr$\rightarrow$Rw & Rw$\rightarrow$Ar & Rw$\rightarrow$Cl & Rw$\rightarrow$Pr & Avg.\\
         \midrule
         ResNet-50~\cite{He2016DeepRL} & \xmark & 34.9 & 50.0 & 58.0 & 37.4 & 41.9 & 46.2 & 38.5 & 31.2 & 60.4 & 53.9 & 41.2 & 59.9 & 46.1 \\
         MCD~\cite{saito2018maximum} & \xmark & 48.9 & 68.3 & 74.6 & 61.3 & 67.6 & 68.8 & 57.0 & 47.1 & 75.1 & 69.1 & 52.2 & 79.6 & 64.1 \\
         CDAN~\cite{long2018conditional} & \xmark & 50.7 & 70.6 & 76.0 & 57.6 & 70.0 & 70.0 & 57.4 & 50.9 & 77.3 & 70.9 & 56.7 & 81.6 & 65.8 \\
         MDD~\cite{zhang2019bridging} & \xmark & 54.9 & 73.7 & 77.8 & 60.0 & 71.4 & 71.8 & 61.2 & 53.6 & 78.1 & 72.5 & 60.2 & 82.3 & 68.1 \\
         BNM~\cite{Cui2020TowardsDA} & \xmark & 52.3 & 73.9 & 80.0 & 63.3 & 72.9 & 74.9 & 61.7 & 49.5 & 79.7 & 70.5 & 53.6 & 82.2 & 67.9 \\
         BDG~\cite{yang2020bi} & \xmark & 51.5 & 73.4 & 78.7 & 65.3 & 71.5 & 73.7 & 65.1 & 49.7 & 81.1 & 74.6 & 55.1 & 84.8 & 68.7 \\
         SRDC~\cite{tang2020unsupervised} & \xmark & 52.3 & 76.3 & 81.0 & 69.5 & 76.2 & 78.0 & 68.7 & 53.8 & 81.7 & 76.3 & 57.1 & 85.0 & 71.3 \\
         \midrule
         PrDA~\cite{kim2020progressive} & \cmark & 48.4 & 73.4 & 76.9 & 64.3 & 69.8 & 71.7 & 62.7 & 45.3 & 76.6 & 69.8 & 50.5 & 79.0 & 65.7 \\
         SHOT~\cite{liang2020shot} & \cmark & 56.9 & 78.1 & 81.0 & 67.9 & \textbf{78.4} & \textbf{78.1} & 67.0 & 54.6 & 81.8 & 73.4 & 58.1 & \textbf{84.5} & 71.6 \\
         SHOT~\cite{liang2020shot} & \cmark & 57.5 & 77.9 & 80.3 & 66.5 & 78.3 & 76.6 & 65.8 & 55.7 & 81.7 & 74.0 & 61.2 & 84.2 & \underline{71.6} \\
         BAIT~\cite{Yang2020UnsupervisedDA} & \cmark & 57.4 & 77.5 & \textbf{82.4} & \textbf{68.0} & 77.2 & 75.1 & \textbf{67.1} & 55.5 & \textbf{81.9} & \textbf{73.9} & 59.5 & 84.2 & 71.6 \\
         BAIT~\cite{Yang2020UnsupervisedDA} & \cmark & 52.2 & 71.3 & 72.5 & 59.9 & 70.6 & 69.9 & 60.3 & 53.9 & 78.2 & 68.4 & 58.9 & 80.7 & \underline{66.4} \\
         \midrule
         \ournet~(ours) & \cmark & \textbf{59.3} & \textbf{78.1} & 79.8 & 65.4 & 75.5 & 76.4 & 65.7 & \textbf{58.0} & 81.0 & 72.0 & \textbf{64.4} & 83.3 & \textbf{71.6} \\
         \bottomrule
         \end{tabular}}
    \end{center}
    \caption{\label{tab:office-home_sup}. Classification accuracies (\%) on the Office-Home dataset (ResNet-50). We adopt underline to denote reimplemented results.}
\end{table*}

\section*{C. More Experimental Results} \label{sec:exp}

\paragraph{Comparison with State-of-the-art Methods.}

We verify the effectiveness of our method on the Office-Home dataset. From Table~\ref{tab:office-home_sup}, the results show that: (1) \ournet~outperforms all the conventional unsupervised domain adaptation methods, which needs to use the source data. (2) \ournet~achieve the competitive performance compared with the state-of-the-art source-free UDA methods, \ie SHOT~\cite{liang2020shot} and BAIT~\cite{Yang2020UnsupervisedDA}. Besides, we also provide our reimplemented results of the published source-free UDA methods on VisDA and Office-31 based on their published source codes (See Table~\ref{tab:visda_sup} and Table~\ref{tab:office_sup}).

\begin{table*}[!h]
\setlength\tabcolsep{2pt}
    \begin{center}
    \scalebox{0.82}{
         \begin{tabular}{lcccccccccccccc}
         \toprule
         Method & Source-free & plane & bicycle & bus & car & horse & knife & mcycl & person & plant & sktbrd & train & truck & Per-class\\
         \midrule
         SHOT~\cite{liang2020shot} & \cmark & 92.6 & 81.1 & 80.1 & 58.5 & 89.7 & 86.1 & 81.5 & 77.8 & 89.5 & 84.9 & 84.3 & 49.3 & 79.6\\
         SHOT~\cite{liang2020shot} & \cmark & 88.5 & 85.9 & 77.9 & 49.8 & 90.2 & 90.8 & 82.0 & 79.0 & 88.5 & 84.4 & 85.6 & 50.5 & \underline{79.4}\\
         BAIT~\cite{Yang2020UnsupervisedDA} & \cmark & 93.7 & 83.2 & 84.5 & 65.0 & 92.9 & 95.4 & 88.1 & 80.8 & 90.0 & 89.0 & 84.0 & 45.3 & 82.7 \\
         BAIT~\cite{Yang2020UnsupervisedDA} & \cmark & 93.8 & 75.4 & \textbf{86.1} & 64.0 & \textbf{93.9} & 96.4 & 88.5 & 81.2 & 88.9 & 88.7 & 86.9 & 39.9 & \underline{82.0}\\
        \midrule
        \ournet~(ours, 40 epochs) & \cmark & 94.8 & 83.6 & 79.7 & \textbf{65.1} & 92.5 & 94.7 & \textbf{90.1} & 82.4 & 88.8 & 88.0 & \textbf{88.9} & {60.1} & {84.1} \\
        \ournet~(ours, 400 epochs) & \cmark & \textbf{95.6} & \textbf{89.0} & 75.4 & 64.9 & 91.7 & \textbf{97.5} & 89.7 & \textbf{83.8} & \textbf{93.9} & \textbf{93.4} & 87.7 & \textbf{69.0} & \textbf{86.0} \\
        \bottomrule
        \end{tabular}
    }
    \end{center}
    \caption{\label{tab:visda_sup}Classification accuracies (\%) on large-scale VisDA dataset (ResNet-101). We adopt underline to denote reimplemented results.}
\end{table*}


\begin{table}[!hbt]
\renewcommand\arraystretch{0.5}
\setlength\tabcolsep{20pt}
\begin{center}
 \scalebox{0.85}{
    \begin{tabular}{ccccc}
    \toprule
    \multirow{2}{*}{$\lambda$} &
    \multicolumn{4}{c}{$\beta$}\cr
    \cmidrule(lr){2-5}
    &0.5&0.7&0.9&0.99\cr
    \midrule
    3 & 81.2 & 83.0 & 83.9 & 83.0\cr
    5 & 81.3 & 82.2 & 84.1 & 83.2\cr
    7 & 79.7 & 81.6 & 83.3 & 83.0\cr
    \bottomrule
    \end{tabular}
    }
    \end{center}
    \caption{\label{tab:elrpara}
 Influence of the trade-off parameters $\beta$ and $\lambda$ in terms of per-class accuracy (\%) on \textbf{VisDA}. The value of $\beta$ is chosen from $[0.5, 0.7, 0.9, 0.99]$ and $\lambda$ is chosen from $[3, 5, 7]$. In each experiment, the rest of hyper-parameters are fixed to the values mentioned in the main paper. \mata{We report the results of the model trained on \textbf{VisDA} for 40 epochs.}}
\end{table}

\begin{table}[!hbt]
\setlength\tabcolsep{1pt}
    \begin{center}
    \scalebox{0.7}{
         \begin{tabular}{lcccccccl}
         \toprule
         Method & Source-free & A$\rightarrow$D & A$\rightarrow$W & D$\rightarrow$W & W$\rightarrow$D & D$\rightarrow$A & W$\rightarrow$A & Avg.\\
         \midrule
         SHOT~\cite{liang2020shot} & \cmark & 93.1 & 90.9 & \textbf{98.8} & 99.9 & 74.5 & 74.8 & 88.7 \\
         SHOT~\cite{liang2020shot} & \cmark & 91.4 & 90.0 & 99.1 & 100.0 & 74.8 & 73.6 & \underline{88.2} \\
         BAIT~\cite{Yang2020UnsupervisedDA} & \cmark & 92.0 & \textbf{94.6} & 98.1 & \textbf{100.0} & 74.6 & 75.2 & 89.1 \\
         BAIT~\cite{Yang2020UnsupervisedDA} & \cmark & 91.3 & 87.4 & 97.6 & 99.7 & 71.4 & 67.2 & \underline{85.8} \\
         \midrule
         \ournet~(ours) & \cmark & \textbf{94.4} & 94.1 & 98.4 & 99.8 & \textbf{76.0} & \textbf{76.6} & \textbf{89.9} \\
         \bottomrule
         \end{tabular}
         }
    \end{center}
    \caption{\label{tab:office_sup}Classification accuracies (\%) on the Office-31 dataset (ResNet-50). We adopt underline to denote reimplemented results.}
\end{table}

\paragraph{Influence of Hyper-parameters.}

In this section, we provide more results for the hyper-parameters $\lambda$ and $\beta$ on VisDA. 
As shown in Table~\ref{tab:elrpara}, our method achieves the best performance with the setting $\beta \small{=} 0.9$ and $\lambda \small{=} 5$ on VisDA.

\paragraph{Visualization of Optimization Curve.}

\begin{figure}[!h] 
\vspace{-0.1in}
\centering
\begin{minipage}{0.49\linewidth}
\subfigure[Total loss curve]{
\includegraphics[width=3.9cm]{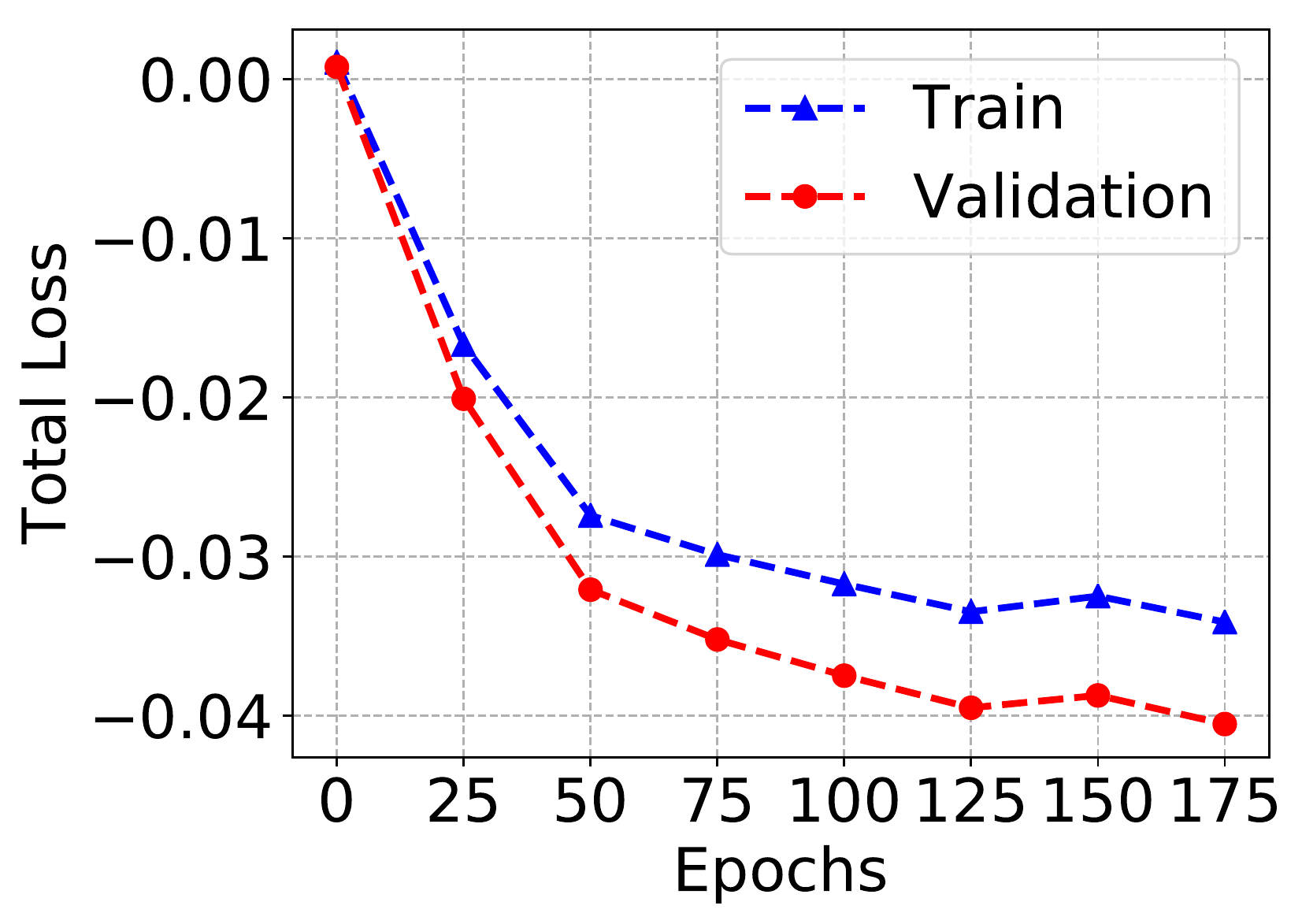}
\label{vis:lce}
}
\end{minipage}
\begin{minipage}{0.49\linewidth}
\subfigure[Accuracy curve]{
\includegraphics[width=3.9cm]{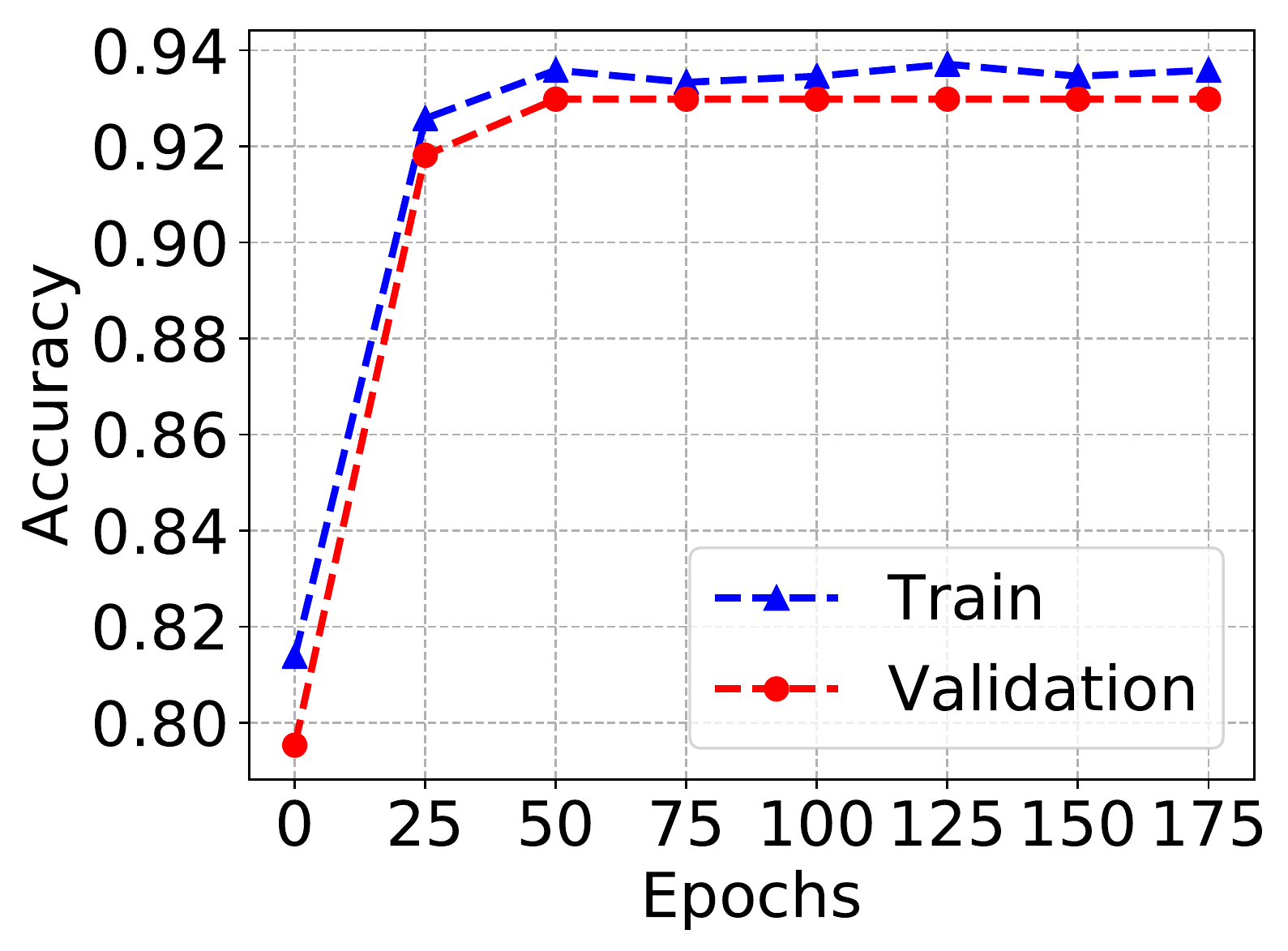}
\label{vis:lcec}
}
\end{minipage}
\label{figdata}
\vspace{-0.1in}
\caption{Optimization curves of \ournet~on \textbf{Office-31}(A$\rightarrow$W).}
\label{vis:curve2}
\vspace{-0.05in}
\end{figure}

Figure~\ref{vis:curve2} shows our  method converges well in terms of the total loss and accuracy in the training  phase. Also, the curve on the validation set means our method does not suffer from pseudo label noise.

\begin{figure}[!h]
		\begin{center}
		\includegraphics[width=4.15cm]{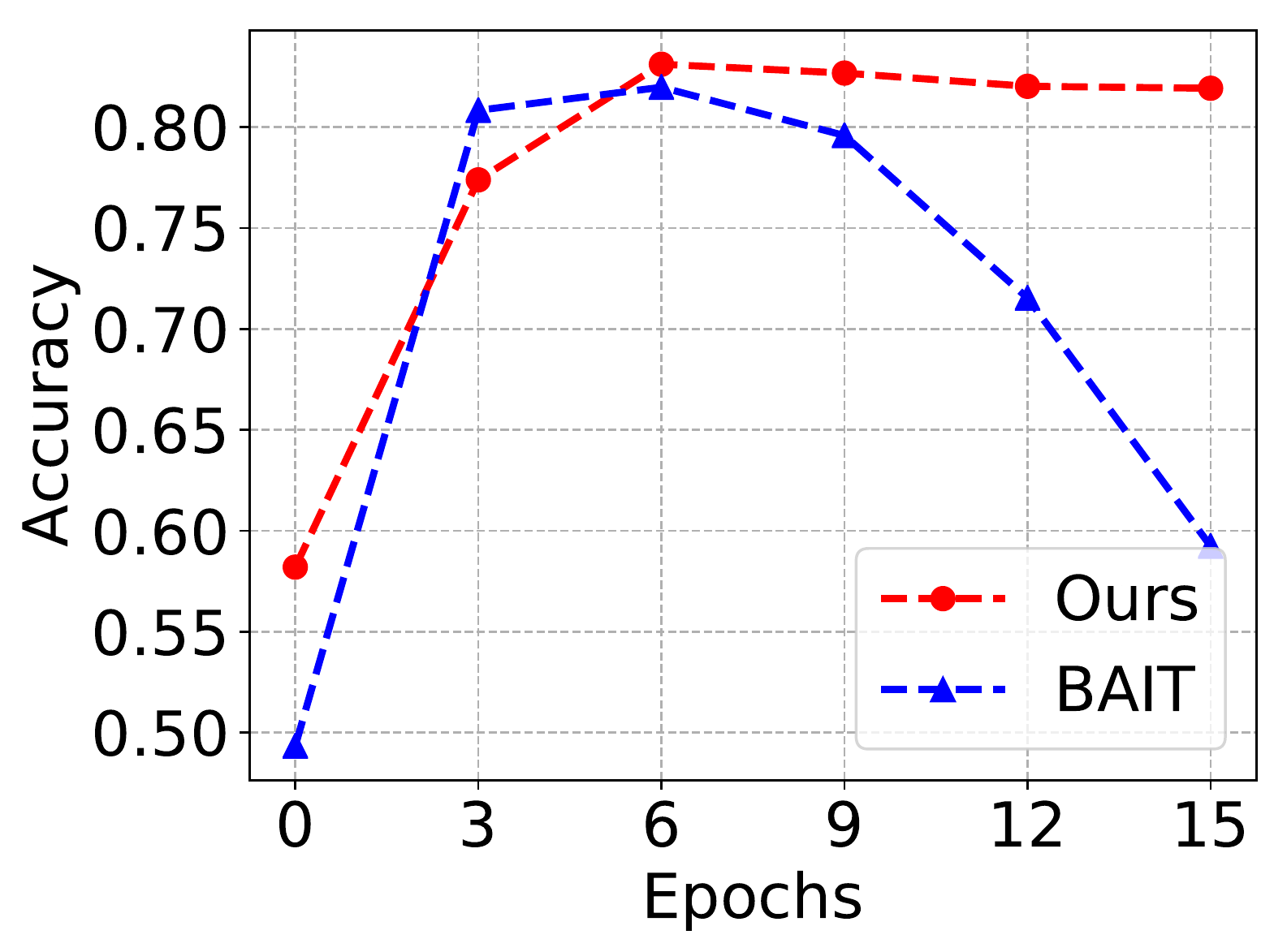}
		\vspace{-0.1in}
		\caption{Testing curves of \ournet~and BAIT on \textbf{VisDA} dataset.}
		\label{fig:noise}
		\end{center}
\vspace{-0.2in}
\end{figure}

\paragraph{Robustness Comparisons with BAIT.} 
As shown in Figure~\ref{fig:noise}, BAIT~\cite{Yang2020UnsupervisedDA} may overfit to mistaken divisions of certain and uncertain sets, leading to poor generalization abilities.
In contrast, our method is more robust and can conquer the issue of pseudo label noise.

\end{document}